\documentclass[conference]{IEEEtran}
\IEEEoverridecommandlockouts
\usepackage{times}
\usepackage{tcolorbox}
\usepackage{amsmath}

\usepackage[utf8]{inputenc}

\usepackage[numbers]{natbib}
\usepackage{multicol}
\usepackage[bookmarks=true]{hyperref}
\usepackage{graphicx}
\usepackage{bbding}
\usepackage{pifont}
\usepackage{hyperref}
\usepackage{cleveref}
\usepackage{multirow}
\usepackage{lineno}
\usepackage{booktabs}
\usepackage{amssymb}
\usepackage{amsfonts}
\usepackage{amsmath}
\usepackage{color}
\usepackage{caption}

\usepackage{algorithm}
\usepackage{algpseudocode}
\usepackage{multirow}

\def\eg{\emph{e.g.}}

\begin{document}


\title{
Demonstrating HumanTHOR: A Simulation Platform and Benchmark for Human-Robot Collaboration in a Shared Workspace
}

\author{\IEEEauthorblockN{Chenxu Wang$^\dag$, Boyuan Du$^\dag$, Jiaxin Xu, Peiyan Li, Di Guo, Huaping Liu$^*$ \thanks{Chenxu Wang, Jiaxin Xu, Peiyan Li and Huaping Liu are with Department of Computer Science and Technology, Tsinghua University, Beijing, China. Boyuan Du is with Fuzhou University, Fujian, China. Di Guo is with the School of Artificial Intelligence, Beijing University of Posts and Telecommunications, Beijing, China. This work was jointly supported by the National Natural Science Fund for Distinguished Young Scholars under Grant 62025304 and National Natural Science Foundation Project under Grant 62273054. $^\dag$ denotes equal contribution. $^*$ denotes the corresponding author: hpliu@tsinghua.edu.cn}}}

\maketitle

\begin{abstract}
Human-robot collaboration (HRC) in a shared workspace has become a common pattern in real-world robot applications and has garnered significant research interest. However, most existing studies for human-in-the-loop (HITL) collaboration with robots in a shared workspace evaluate in either simplified game environments or physical platforms, falling short in limited realistic significance or limited scalability. To support future studies, we build an embodied framework named HumanTHOR, which enables humans to act in the simulation environment through VR devices to support HITL collaborations in a shared workspace. To validate our system, we build a benchmark of everyday tasks and conduct a preliminary user study with two baseline algorithms. The results show that the robot can effectively assist humans in collaboration, demonstrating the significance of HRC. The comparison among different levels of baselines affirms that our system can adequately evaluate robot capabilities and serve as a benchmark for different robot algorithms. The experimental results also indicate that there is still much room in the area and our system can provide a preliminary foundation for future HRC research in a shared workspace. More information about the simulation environment, experiment videos, benchmark descriptions, and additional supplementary materials can be found on the website: https://sites.google.com/view/humanthor/.
\end{abstract}

\section{Introduction}
With the development of various techniques in multi-modal perception, reasoning, planning, and control, robots have become increasingly powerful and are gradually deployed in a variety of application scenarios. Human-robot collaboration (HRC) has been gaining more and more interest and become a hot-spot problem in various real-world domains, such as industry \cite{villani2018survey, kumar2020survey}, surgery \cite{kaplan2016toward, long2023human}, rescue \cite{murphy2004human},  health caring \cite{broadbent2011human}, agriculture \cite{vasconez2019human}, and home service \cite{lee2005designing}. 

A common and realistic collaboration scenario is the loosely coupled collaboration in a shared workspace, such as doing housework, where people act independently in the house towards a shared goal and may communicate with the robot to exchange information. Studying HRC in such a scenario necessitates two key features of the simulation system: (1) The task is human-in-the-loop (HITL), requiring the environment to support real-time interaction with humans. (2) The human and the robot work in a shared workspace, where they share the same level of observation and similar capabilities rather than the instructor-follower paradigm in previous works \cite{gao2023alexa, padmakumar2022teach}. For example, for the task of putting an apple in the fridge, the human has to first search for the apple and then search for the fridge to put the apple in it. With the human-robot collaboration, the human and robot can simultaneously search for the apple and fridge respectively in the shared workspace. When the robot finds the fridge, it could report the position of the fridge to the human and the human could take the apple directly to the fridge, improving working efficiency.

\begin{figure*}[thbp]
    \centering

    \includegraphics[width=\textwidth]{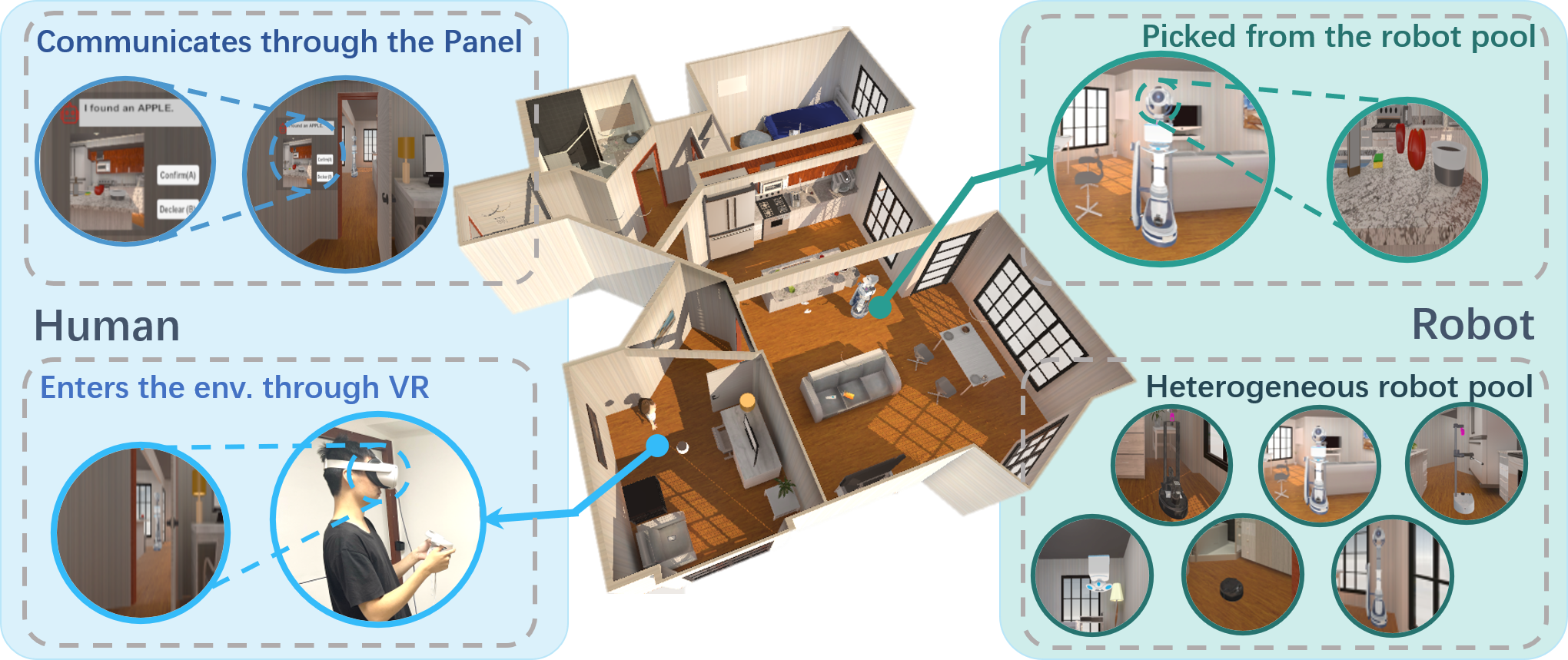}
    \caption{An overview of the HumanTHOR system, where the human can act in the simulator through the VR device with the first-person view akin to the robot. The system also supports the top-down view with instant displaying of the positions and orientations of the human and the robot.}
    
   \label{fig: demo}
\end{figure*}

To support further study, establishing a system along with corresponding benchmarks is crucial. 
However, existing systems may not be sufficient for studying the HITL collaboration with robots in a shared workspace due to various limitations.
One mainstream practice is to study HRC on physical systems, such as performing collaborative assembly of furniture \cite{maccio2022mixed}, pick and place tasks \cite{ferrari2022bidirectional}, and simple picking objects \cite{rosen2020mixed}. 
Despite the practical effectiveness, physical systems may suffer from high economic costs, limited reproducibility, and relatively small scales.
On the other hand, studying cooperation in 2D games such as the Overcooked \cite{carroll2019utility} and the cooperative table-carrying game \cite{ng2023takes} may fall short in being too simple to generalize to physical robots, despite their reproducibility.


Fortunately, recently there emerge many embodied simulators such as AI2THOR series \cite{kolve2017ai2, deitke2020robothor, ehsani2021manipulathor}, which appear to be excellent platforms for studying robot algorithms, for both being able to scale up and approximate the real world.
However, although many studies have noticed the importance of human factors and extended the benchmarks to incorporate various modalities of human collected data, such as \cite{gao2022dialfred, wu2021communicative, gao2023alexa}, there have been limited attempts to study the HITL collaboration in such embodied simulators. 
 
To support the study of HITL collaboration with robots in a shared workspace, we build an embodied human-robot collaboration environment based on the AI2THOR framework, namely HumanTHOR. As shown in \Cref{fig: demo}, humans can immersively control their avatars to cooperate with the robot in the environment through VR devices. Compared to previous works that use VR to collect human demonstration \cite{li2022igibson, FuRSS23,li2024sensor}, our environment is dedicated to HRC tasks with various functional supports, such as enabling synchronous collaboration between humans and robots and providing image-text communication interfaces. 

To demonstrate our system, we further implement a preliminary HRC benchmark of everyday tasks, including two representative tasks: \textit{object goal navigation} and \textit{mobile manipulation}. To generate realistic and diversified episodes, the initial states of object arrangement are sampled from a set of scene priors. To validate our system, we conduct a user study by employing a rule-based robot and an oracle robot as baselines. Experimental results show that robots can significantly assist humans in everyday tasks, and our system serves as an effective testbed for HRC studies.

We summarize our main contributions as follows: 
\begin{itemize}
\item We develop the HumanTHOR system, which enables real-time human-robot collaboration in shared workspace with multi-modal communication through VR devices. 
\item We implement a benchmark of everyday tasks, which can be used for studying HRC in embodied scenarios with a human-robot communication mechanism.
\item We run a user study with various settings of robots. The results suggest the effectiveness of our system to evaluate the capability of robot algorithms and further serve as a testbed for studying HRC problems.
\end{itemize}

This paper is organized as follows: in \Cref{sec: RelatedWork}, we review recent works that are related to our study. We introduce the architecture, main characteristics, and implementation of the HumanTHOR system in \Cref{sec: System}. In \Cref{sec: Benchmark}, we elaborate on the designation and details of the HRC benchmark. We then present the experiment results and corresponding analysis in \Cref{sec: Experiments}. In \Cref{sec: Extension}, we demonstrate the extensibility of our system and extensions regarding multi robots with more complex HRC tasks and robot algorithms. Finally, we conclude and discuss the future work in \Cref{sec: Conclusions}. 

\section{Related Work}
\label{sec: RelatedWork}

\subsection{Human-Robot Collaboration}
The pursuit of creating collaborative and user-friendly robots has attracted substantial attention in the research community. To provide an immersive user experience, Virtual Reality (VR) has been frequently utilized as the user interface in human-robot interaction studies \cite{dianatfar2021review}. Beyond serving as a display media \cite{promsutipong2022immersive}, the matched controllers enable humans to teleoperate robots or their avatars \cite{de2021leveraging, kohn2018towards, stotko2019vr}. Such VR-based controls are also extensively used for collecting human demonstrations for robot learning \cite{li2022igibson, fu2023demonstrating}.  

Since VR provides virtual and digital visions, such techniques can also be integrated with real-world systems and become Augmented Reality (AR) or Mixed Reality (MR), which significantly benefit collaborations in the physical world. MR can serve as a communication medium to convey the intention of robots \cite{rosen2020mixed, maccio2022mixed}, while AR can facilitate the study of human-robot collaboration in shared workspace \cite{qiu2020human}.

To provide immersive environment exploration and operation and approximate the human-robot collaboration in the real world, we employ VR as the user interface in our HumanTHOR system.

\subsection{Simulators and Benchmarks for Embodied Intelligence}
Along with the rising interest in embodied intelligence, a considerable number of simulation environments emerge. Environments such as AI2THOR series \cite{kolve2017ai2, ehsani2021manipulathor, deitke2020robothor}, iGibson \cite{shen2021igibson, li2022igibson}, Habitat \cite{savva2019habitat, szot2021habitat}, Matterport3D \cite{chang2017matterport3d}, and RFUniverse \cite{fu2023demonstrating} have present vivid simulation with various physics systems and make it possible for building benchmark for tasks such as navigation or manipulation. 

Beyond the platforms, various benchmarks for embodied tasks have been proposed, such as question answering \cite{das2018embodied}, visual language navigation \cite{anderson2018vision}, scene graph generation \cite{li2022embodied}, and performing composite everyday tasks \cite{shridhar2020alfred}. After achieving adorable success in these tasks, researchers have turned their gaze to multi-agent settings \cite{liu2022multi, liu2022embodied}, considering human-factors \cite{silva2023online, zhang2023human}, and collaboration with humans. Beyond a single paragraph of instructions, GesTHOR \cite{wu2021communicative} augment the embodied navigation with human gesture indications. HandMeThat \cite{wan2022handmethat} takes one more step toward HRC by playing a record of human actions and instructions and then trains the robots to follow such instructions. Recent benchmarks consider communication between humans and robots. DialFRED \cite{gao2022dialfred} and TEACh \cite{padmakumar2022teach} extend the instructions into dialogue form and encourage the robots to actively raise questions. Alexa Arena \cite{gao2023alexa} builds an interactive environment where the instructor and executor can interactively communicate with natural language. While the benchmark is built on offline data, the environment may also support HITL collaborations. 

To comprehensively support our intended HRC study, the system requires several key features, including: (1) \textbf{Vivid 3D simulation}, for both aligning the simulation to real-life robot application and better human experience; (2) \textbf{HITL interaction}, for conducting experiments with real humans instead of with proxy agents or offline data; (3) \textbf{Shared workspace collaboration}, which is the domain we intend to study; and (4) \textbf{Instant communication}, a vital component for human-robot interaction and simulating the collaboration in the real world. For loosely coupled collaboration where the human and robots may not always stay in the vision of each other, the system is expected to support multimedia message-style communication, including both image and text.
Additionally, the system is expected to have a VR interface for both immersive human experience and better aligning human activities to the real life.

However, existing simulation environments are not sufficient enough to simultaneously meet all the requirements mentioned above. To overcome this shortage, we build the HumanTHOR system that includes all the abovementioned features. A detailed comparison between the proposed HumanTHOR and the existing simulation environment is delineated in \Cref{tab: SystemComparison}.

\begin{table*}
    \centering
    \begin{tabular}{|c|c|c|c|c|c|c|} 
    \hline
    \multirow{2}{*}{Benchmark} & \multirow{2}{*}{Simulation Domain} & \multirow{2}{*}{VR Interface} & \multirow{2}{*}{HITL Collaboration} & \multirow{2}{*}{Shared Workspace} & \multicolumn{2}{c|}{Communication} \\  
    \cline{6-7}
     & &  &  &  & Image & Text \\ 
    \hline
    Overcooked \cite{carroll2019utility} & 2D & \ding{55} & \ding{51} & \ding{51} & \ding{55} & \ding{55} \\ 
    It takes two \cite{ng2023takes} & 2D & \ding{55}  & \ding{51} & \ding{51} & \ding{55} & \ding{55} \\ 
    HandOverSim \cite{chao2022handoversim} & 3D & \ding{55} & \ding{55} & \ding{51} & \ding{55} & \ding{55}  \\ 
    VR Kitchen \cite{gao2019vrkitchen} & 3D  & \ding{51} & \ding{55} & \ding{55} & \ding{55} & \ding{55} \\ 
    DialFred \cite{gao2022dialfred} & 3D & \ding{55} & \ding{55} & \ding{55} & \ding{55} & \ding{51}   \\ 
    TeaCh \cite{padmakumar2022teach} & 3D & \ding{55} & \ding{55} & \ding{55} & \ding{55} & \ding{51}   \\ `
    HandMeThat \cite{wan2022handmethat} & 3D & \ding{55} & \ding{55} & \ding{55} & \ding{55} & \ding{51}   \\ 
    GesTHOR \cite{wu2021communicative} & 3D & \ding{51} & \ding{55} & \ding{55} & \ding{55} & \ding{55} \\ 
    Alexa Arena \cite{gao2023alexa} & 3D & \ding{55} & \ding{51} & \ding{55} & \ding{55} & \ding{51} \\ 
    \hline
    HumanTHOR & 3D & \ding{51} & \ding{51} & \ding{51} & \ding{51} & \ding{51}  \\
    \hline
    \end{tabular}
    \caption{Comparison between HumanTHOR and related HRC benchmarks across various aspects.}
    \label{tab: SystemComparison}
\end{table*}

\begin{figure}[t]
    \centering
    \includegraphics[width=0.45\textwidth]{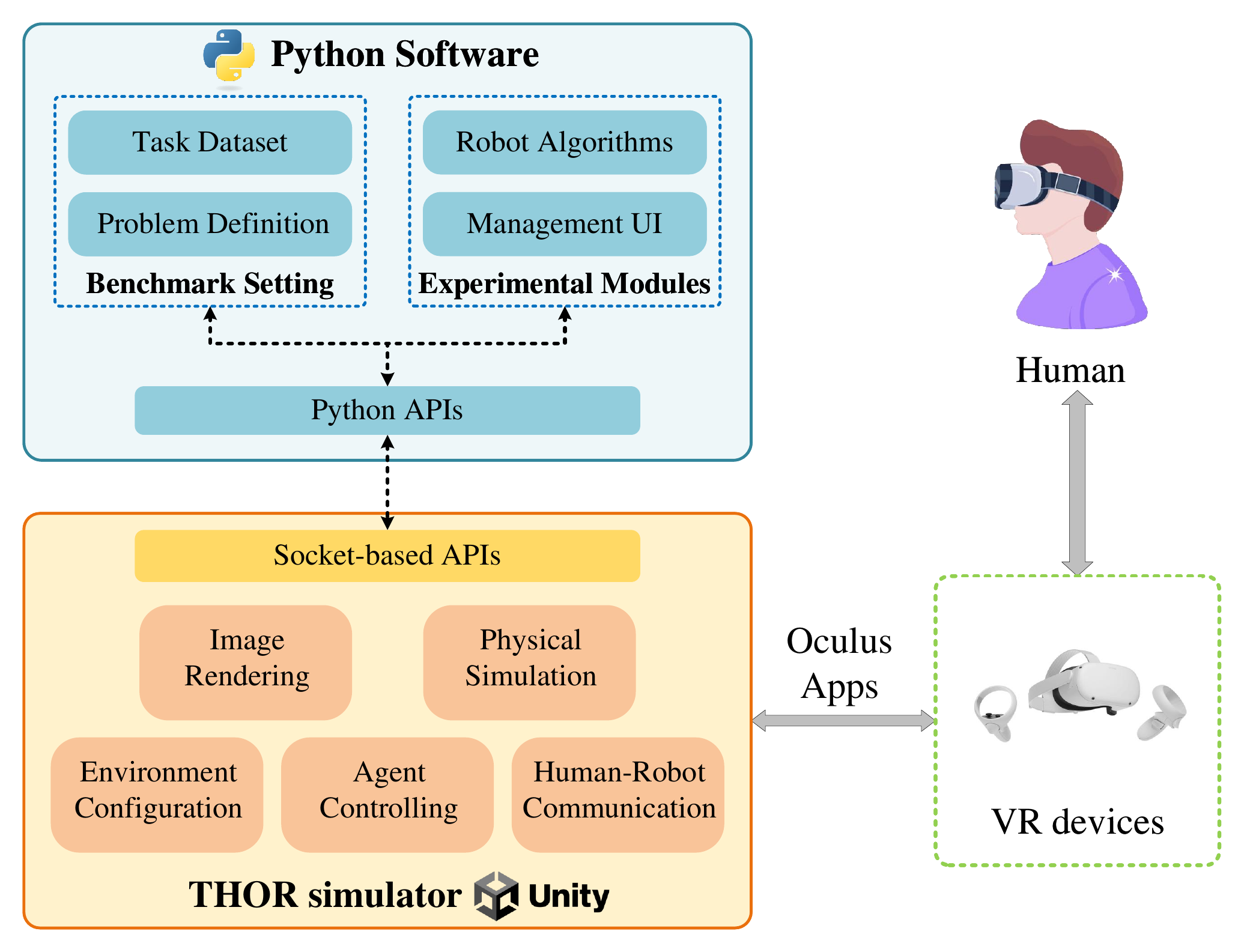}

    \caption{The architecture of our HumanTHOR system.}
    \label{fig: SystemArchitecture}
\end{figure}

\section{System Framework}
\label{sec: System}

\subsection{System Overview}
Aiming to support human-in-the-loop collaborations in a shared workspace with flexible and multimodal communication, we design and implement the HumanTHOR system based on the AI2THOR platform. As the hierarchical architecture illustrated in \Cref{fig: SystemArchitecture}, our system consists of three components: 
\begin{itemize}
    \item The \textbf{THOR simulator} implemented with the Unity framework based on the AI2THOR infrastructure, responsible for physical simulation, scene rendering, and solving the interaction between agents and environments. The simulator exposes a set of socket-based APIs for controlling.
    \item The \textbf{python software} that supports further HRC study, including extensible benchmarks which will be introduced in \Cref{sec: Benchmark}, a code framework for running robot algorithms and a web user interface for controlling and monitoring the HRC experiments.
    \item \textbf{VR devices}. We use Meta Quest 2 in our system, including a headset and a pair of controllers. VR devices are connected to our simulator through the Oculus platform.
\end{itemize}

By integrating VR devices into our system, humans and robots can concurrently work within the environment, thereby achieving HITL and human-robot collaboration in a shared workspace.

\subsection{Supported APIs}
\label{sec: SystemAPI}
To support various requirements of HRC benchmarks, our system provides dozens of APIs in the HumanTHOR environment, which are generally in the following four categories:
\begin{itemize}
    \item \textbf{Environment configuration}: We provide a series of APIs for setting up the environment, including \textit{selecting scenes}, \textit{initialization}, and \textit{object settings}. To support the development of benchmarks to set up customized tasks, environmental information is also obtainable, such as the amount, positions, and states of objects. 
    \item \textbf{Agent control}: Similar to previous works, HumanTHOR supports agent actions such as \textit{move}, \textit{rotate}, \textit{teleport}, \textit{pick}, and \textit{place}. For the convenience of customizing benchmarks, we use a flexible format in which the \textit{move} action and the \textit{rotate} action take a vector that indicates the variation as input, and leave the designation of atomic actions to the benchmark level. 
    \item \textbf{Perception and monitoring}: Following the conventions, our environment provides an ego-centric RGB observation and a depth map. We also provide a top-view image and the positions of all agents for monitoring.  
    \item \textbf{Communication}: A major contribution of our HumanTHOR environment is the multimodal communication module for collaboration. We provide communication APIs including sending image-text messages, responding to messages, and querying the response status.   
\end{itemize}

\subsection{User Interface and VR Supports}
\begin{figure}[t]
\centering
\includegraphics[width=3.4in]{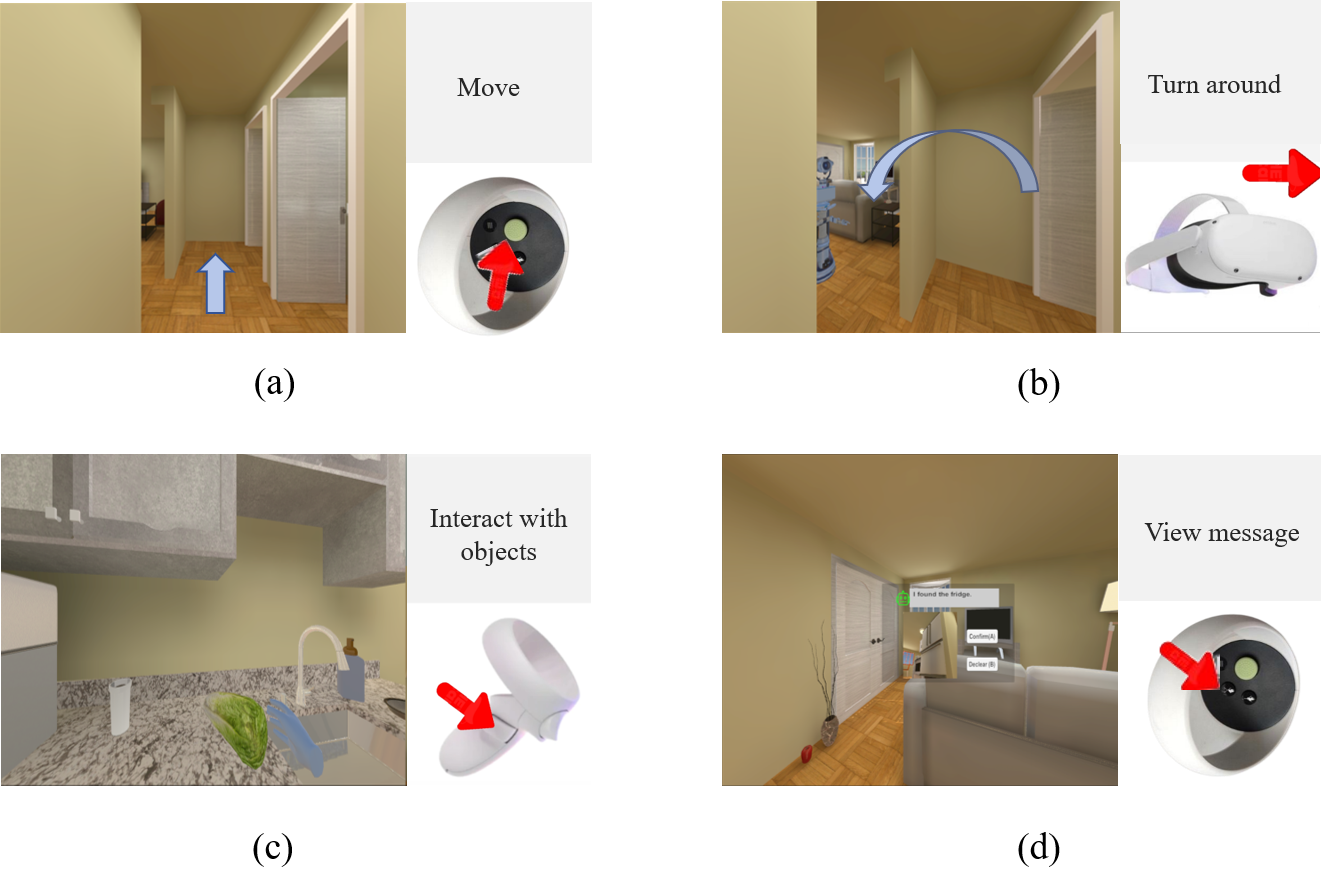}
\caption{Interacting with the environment with VR devices. (a) Moving the human avatar by operating the joystick on the VR controller. (b) With the help of the sensors on the head-mounted display, humans can conveniently rotate the angle of view by turning around in reality. (c) When being close enough to an object, humans can pick up a movable object by pressing the side button as shown in the figure, or open a receptacle such as a fridge. (d) After receiving a message from the robot, humans can make quick responses with the A/B buttons on the controller. In our benchmark of collaborative tasks, the message is presented in a dialogue box in the human view, where button A is for confirmation and button B is for decline. After confirmation, a map with the relative positions of the robot and human will be displayed, which can also be hidden or redisplayed by the button operation.}
   \label{fig: vr operation}
\end{figure}

To provide an immersive experience for human players, we employ VR as the user interface. Humans can perceive the environment from the first perspective with the headset, and control their avatars through the buttons and the rocking bars on the controllers, which enable them to move in the environment and interact with objects. We further develop several interfaces for human-robot communication including an image-text message box and a location map, all can be presented in the human vision. We present these operations in \Cref{fig: vr operation}. 

\section{Benchmarking HRC with everyday tasks}
\label{sec: Benchmark}

The HumanTHOR platform is eligible for studying various HRC tasks at various levels, being available for both evaluating robot algorithms and human-robot collaboration studies. 
We model the benchmark structure in a hierarchical way as illustrated in \Cref{fig: BenchmarkArch}, where we start by supporting elemental tasks, based on which we can perform practical everyday tasks, such as visual language navigation \cite{shridhar2020alfred}, room rearrangement \cite{batra2020rearrangement, weihs2021visual}, tidying up \cite{sarch2022tidee}.
Ultimately, HumanTHOR supports studying high-level scientific problems such as Theory-of-Mind and human trust modeling.

To preliminarily validate the effectiveness of our system, we build benchmarks for the elementary \textbf{object goal navigation} task and a compound HRC task, \textbf{mobile manipulation}, as defined in the following:

\begin{figure}[t]
   \centering
    \includegraphics[width=0.45\textwidth]{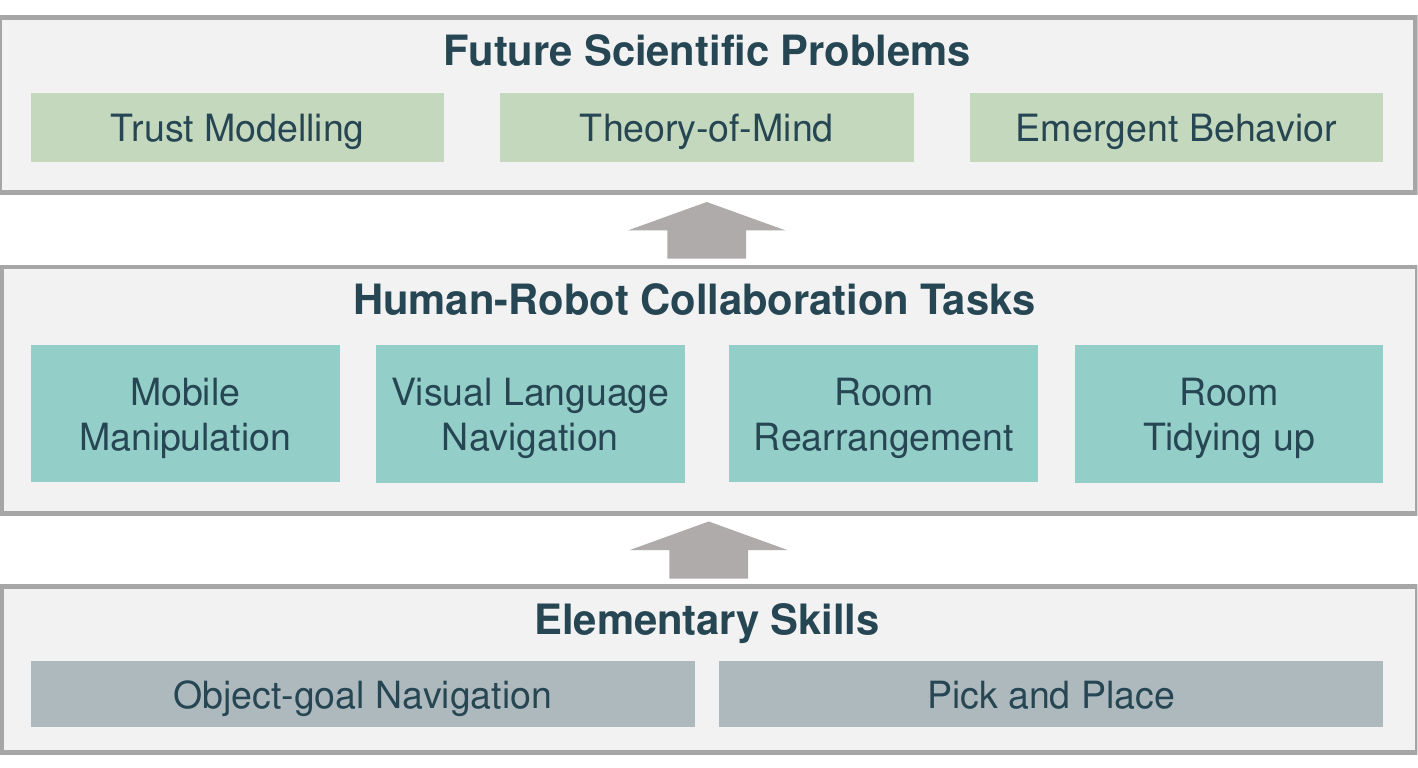}

   \caption{The hierarchical benchmarks supported by the HumanTHOR platform.}
   \label{fig: BenchmarkArch}
\end{figure}

\begin{itemize}
    \item \textbf{Object goal navigation}, a representative and fundamental task in the embodied intelligence domain. Such tasks target a common situation in which a human is searching for something in the house. The task succeeds when the human successfully finds the target object. 

    \item \textbf{Mobile manipulation}. In this task, participants need to pick up the target object and place it at the destination place. This is a fundamental type of everyday task, which is also adopted by ALFRED \cite{shridhar2020alfred}, being more complex than navigation tasks since it involves multiple objects and interactions with the environment.
\end{itemize}

\begin{figure}[t]
   \centering
    \includegraphics[width=3.4in]{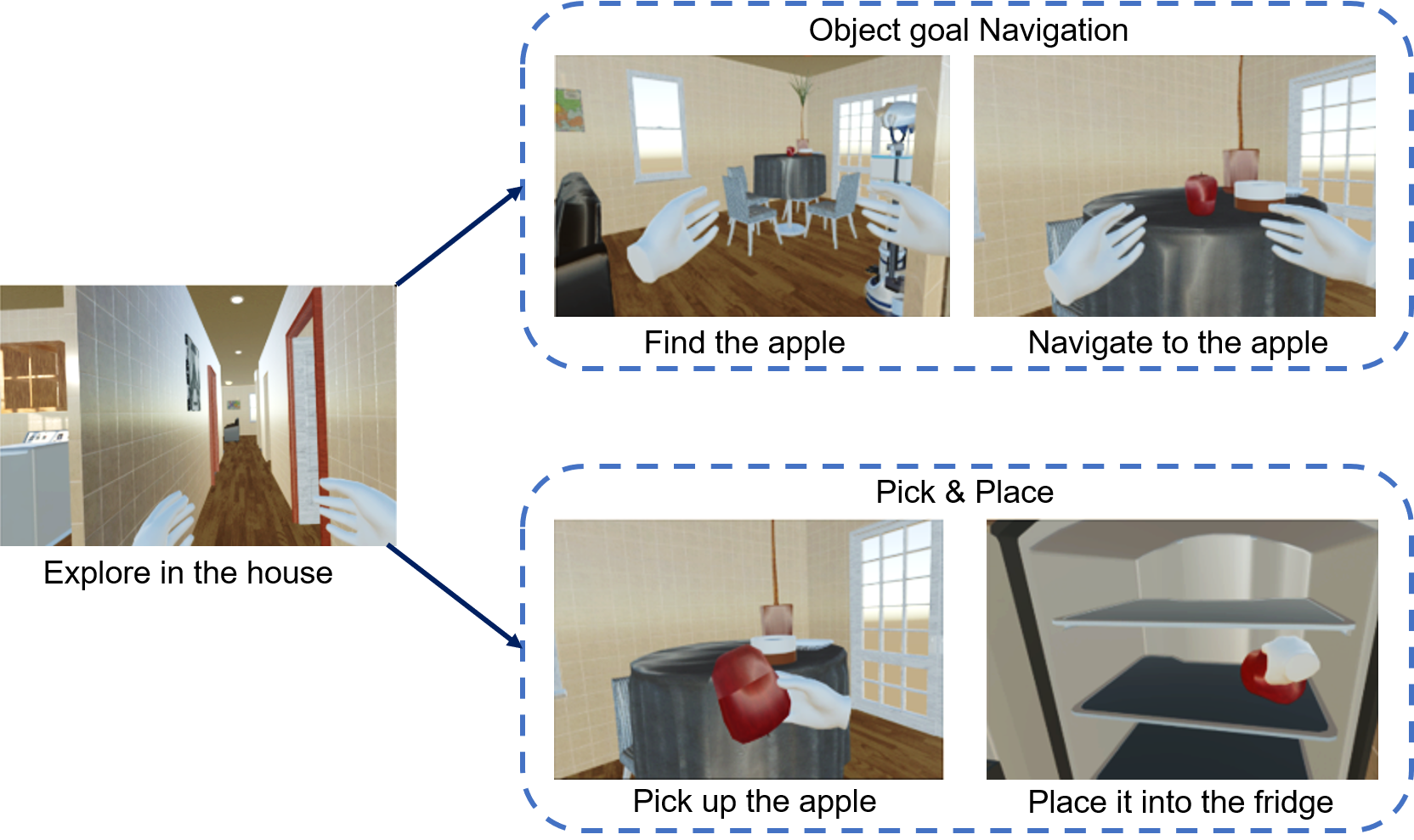}
   
   \caption{The general processes of navigation tasks and manipulation tasks. The mobile manipulation task is more complex and difficult since it requires further manipulation after successfully finding the target.}
   \label{fig: TaskDemo}
\end{figure}

Two examples of tasks are presented in \Cref{fig: TaskDemo}. Similar to previous works \cite{shridhar2020alfred, gao2023alexa}, our tasks are defined with three essential components: \textbf{initial state} that determines the initial arrangement of objects, \textbf{goal definition} which formalizes the criterion of success, and \textbf{auxiliary information} such as natural language description of the tasks.

\subsection{Object-centric Interactive Communication}

\begin{figure}[t]
    \centering
    \includegraphics[width=0.5\textwidth]{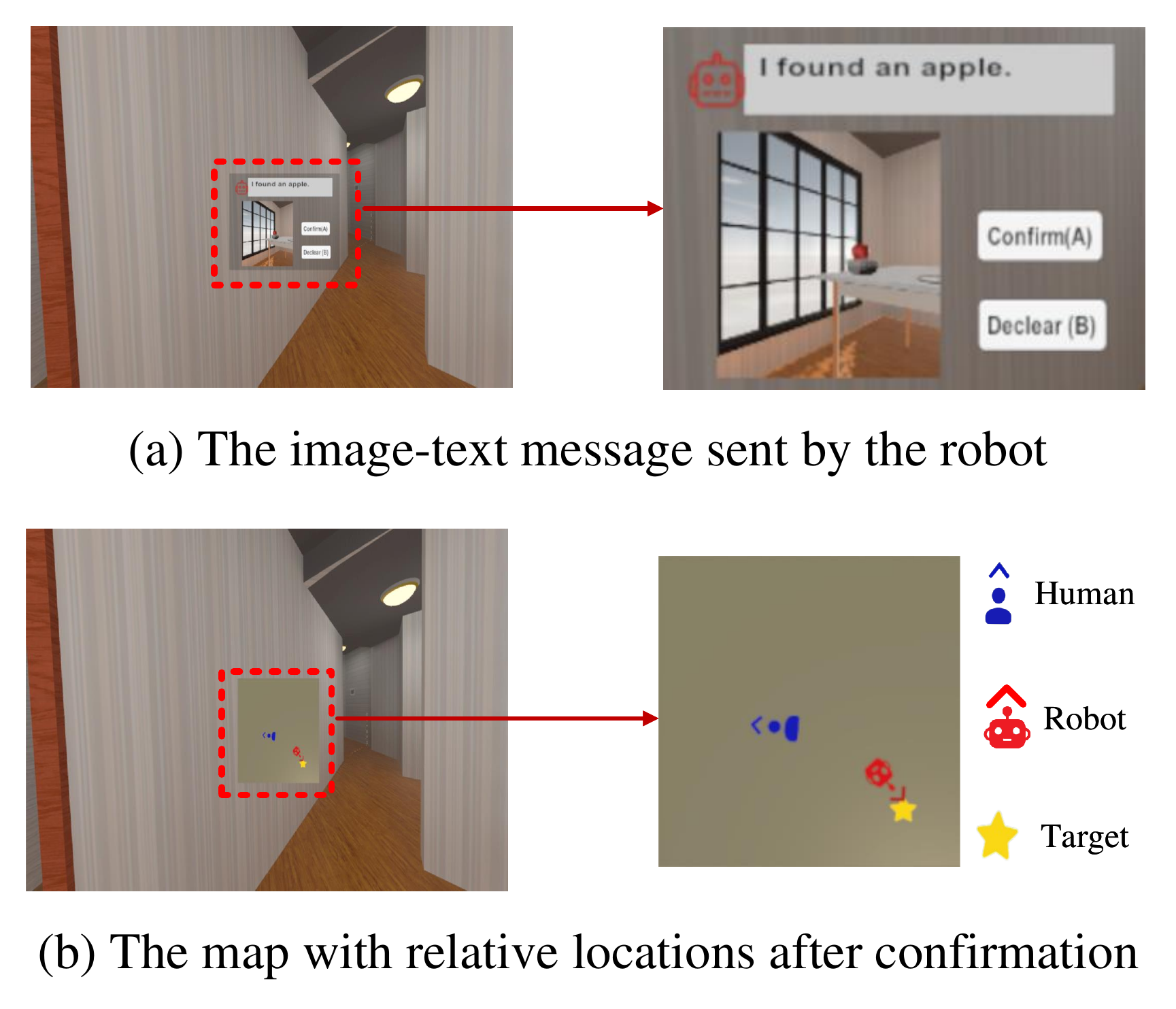}

    \caption{An illustration of the object-centric communication for human-robot collaboration.}
    \label{fig: communication}
\end{figure}

We implement an object-centric communication flow with corresponding user interfaces for human-robot collaboration in everyday tasks. Whenever the robot feels necessary to communicate with object-related information, it can send an image-text message as shown in \Cref{fig: communication} (a), which consists of the observation of the robot and a custom message text, as in \Cref{fig: communication} (b). If humans are interested in the message, they can press the button to confirm. Then a map with relative positions will be shown subsequently to help the human go to the suggested place. Both user interfaces can be temporally hidden and displayed whenever the human wants. We also provide an API for the robot to query whether its suggestion is accepted or declined. 

It is worth noting that robots are free to choose when and what to send in the message, and are allowed to cover the previous message with new ones. Combined with hierarchical levels of capabilities in heterogeneous teamwork, there still exists a large room for further study. For complex tasks, robots may need to carefully estimate human mental state to choose the appropriate time for sending messages, and conveying important information while not bothering humans too much.

\subsection{Generating Everyday Tasks with Scene Priors}
Our benchmark incorporates 10 scenes and 14 objects, including 8 receptacle objects and 6 target objects. We obtain 19 mobile manipulation task templates by artificial annotations, which are in the form of \textit{Picking a/an \{target object\} and place it in/on a/an \{receptacle objects\}}. 

\begin{figure}[t]
    \centering
    \includegraphics[width=3.4in]{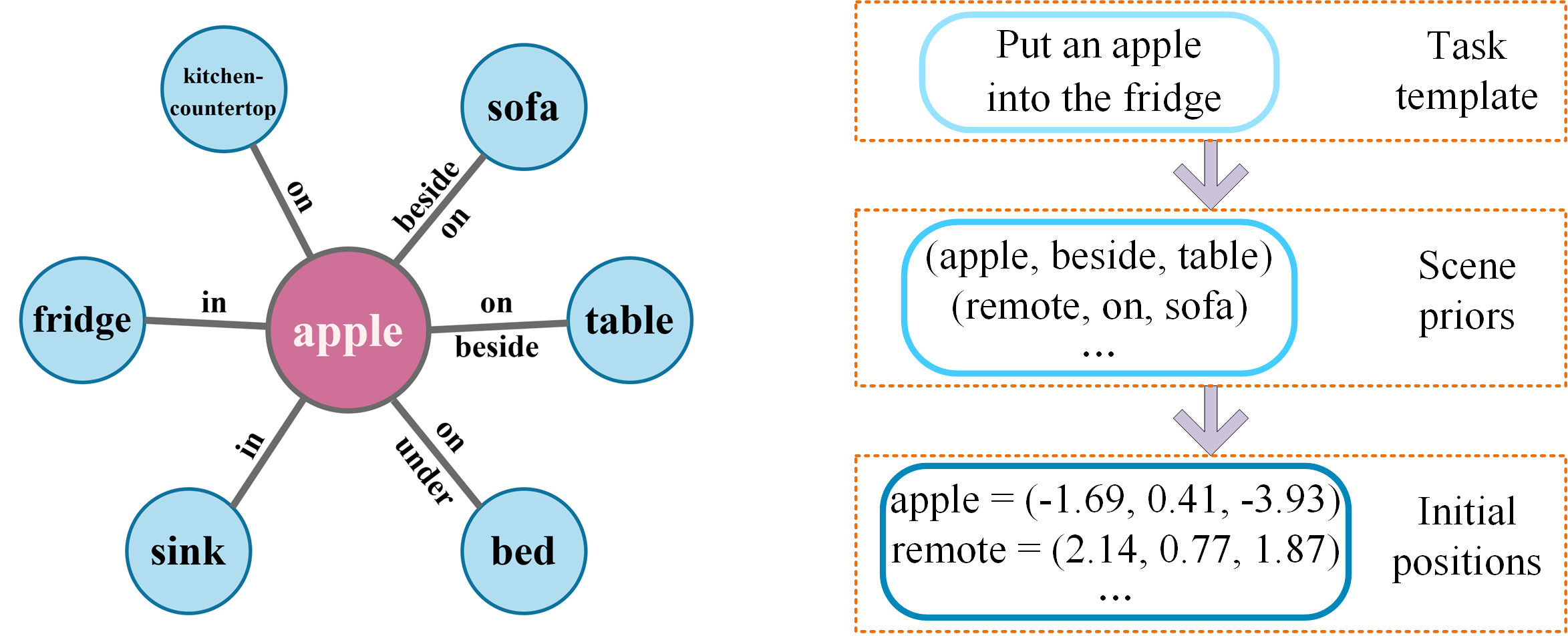}
    
    \caption{An example of scene priors and the task generation pipeline.}
    \label{fig: TaskGeneration}
\end{figure}

We generate the initial state with the guide of scene priors like exhibited in \Cref{fig: TaskGeneration}. We first build a knowledge graph with 65 triplets by human annotations, each triplet represents a possible prior relation $(target, relation, reference)$. We weight the triplets by combining the similarities between the textual features of objects adopted from the pretrained BERT model \cite{devlin2018bert} and human priors.

We take three steps to instantiate the task as illustrated in \Cref{fig: TaskGeneration}. First, we pick a task template that is applicable to the scene to determine the \textit{goal}. Then we construct a scene graph by sampling scene priors from the knowledge graph. Each sampled relation corresponds to an area where the object should be placed, and the specific position of each object is subsequently determined.
Following this program, we generate 565 navigation tasks and 1583 mobile manipulation tasks. All initial relations of the target object are sampled once for each task template in each scene, while the irrelevant objects are not limited.

\section{Experiments}
\label{sec: Experiments}
To validate our proposed HumanTHOR system, we conduct a preliminary user study on a subset of our built benchmark, focusing on the mobile manipulation tasks, which are more complex and encompass the navigation task. We aim to answer two questions in the experiment: (1) the necessity of HRC: can the robot help humans perform the tasks? (2) The effectiveness of our system as an evaluation benchmark: can our experiment distinguish different levels of robots and evaluate their capabilities?

\subsection{Baselines}

We incorporate two simple robot baselines in the experiments. For both robots, the enabled actions include \textit{move}, \textit{rotate}, and \textit{send message}. The atomic movement distance can not exceed 0.5 meters, the angle of pitch is limited to $[ -30^\circ, 30^\circ]$, and sending a message requires a custom text and an estimated position. The details of the baselines are as follows:

\begin{itemize}
\item Frontier agent. It is a rule-based agent that explores the environment with the frontier algorithm and sends messages whenever it detects the target. To simplify the task, we endow the frontier agent with an ideal object detector that can get ground truth object detection within 1.5 meters. 
\item Oracle. The oracle robot knows the positions of all objects and will navigate to the target object along the shortest path. However, it still follows the communication framework and only sends messages when it is close enough to detect the target.
\end{itemize}

Both baselines take a default setting to search for the target object instead of the receptacle or dynamic task allocation. Though the baselines may not be practical or optimal, we clarify that they are used for the verification and calibration of our system, where the oracle serves as an approximation of the upper bound and the frontier serves as a rule-based baseline without training or intelligent algorithms.
Besides, we also set up a control group that has no robot. 

\begin{table}[t]
    \centering
    \begin{tabular}{|c|c|c|c|}
        \hline
        Robot setting & SR (\%) & TWSR (\%) & Adoption Rate (\%) \\
        \hline
        No Robot & 63.8 & 46.7 & / \\
        Frontier & 72.9 & 54.3 & 56.0 \\
        Oracle & 91.4 & 66.7 & 90.0 \\
        \hline
    \end{tabular}
    \caption{Average success rate, time-weighted success rate, and adoption rate in all three settings.}
    \label{tab: QuantResult}
\end{table}

\begin{figure}[t]
\centering
\begin{minipage}[t]{0.23\textwidth}
\centering
\includegraphics[width=3.8cm]{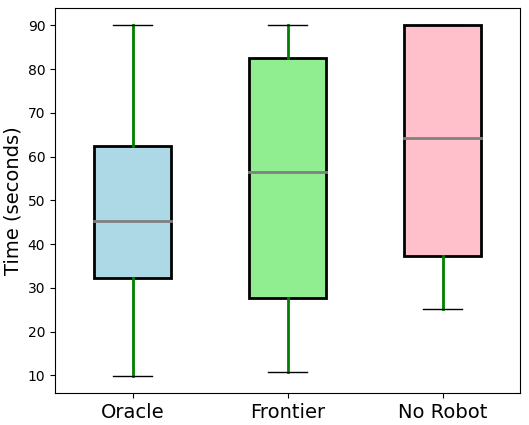}
\caption{Average time spent to complete the tasks}
\label{fig: TimeSpent}
\end{minipage}
\begin{minipage}[t]{0.23\textwidth}
\centering
\includegraphics[width=3.8cm]{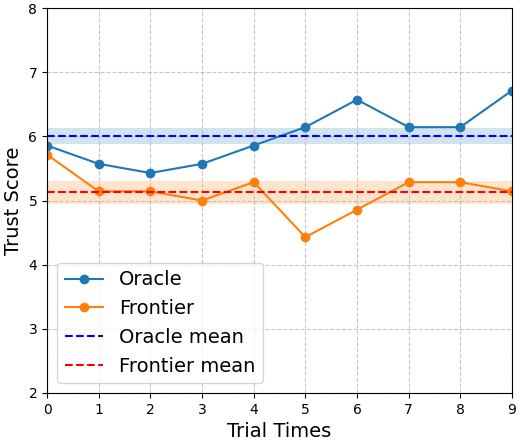}
\caption{The variation of average human trust scores over trials.}

\label{fig: HumanTrust}
\end{minipage}
\end{figure}

\begin{figure*}[t]
    \centering
    \includegraphics[width=\textwidth]{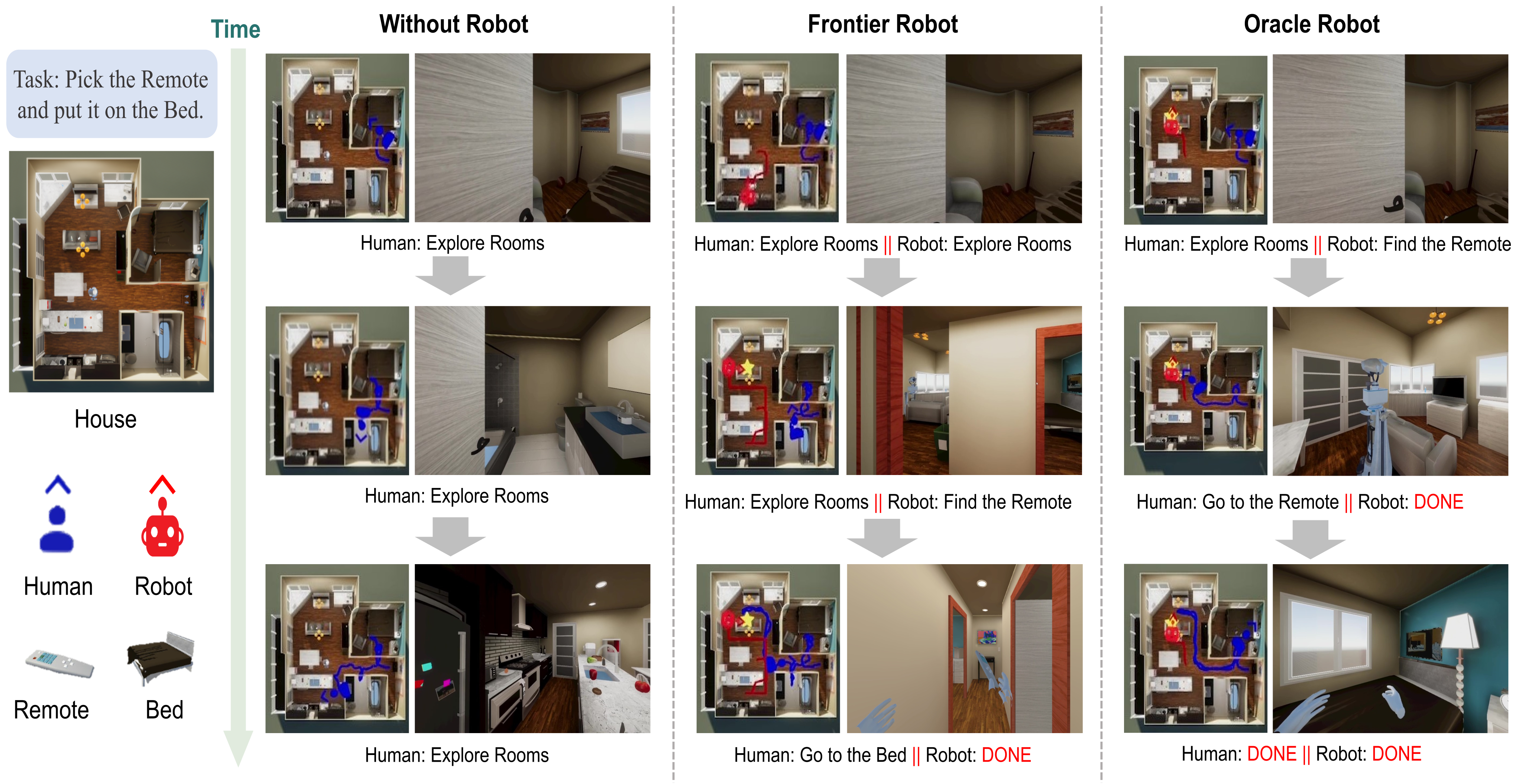}

    \caption{A representative task for demonstrating the effect of robot assistants. We take one case in each robot setting and present the top-down maps and human observations at several keyframes. The colored lines in the top-down maps denote the trajectories of the robot and the human.}
    \label{fig: CaseStudy}
\end{figure*}

\subsection{Experimental setup}
We sample 3 mobile manipulation tasks in each scene for the experiments, in total 30 tasks in all 10 scenes. We invite 18 human participants, which are equally divided into 3 groups. Each group plays within a robot setting for 10 random tasks, one task in each scene. For a fair comparison, the sequence of tasks is random for each participant. For each task, the participant has 90 seconds to finish the task. After finishing each task, the participant is asked to give a trust score for the robot (if applicable) to study human trust. The trust score is an integer from 1 to 7, where 1 denotes not trust at all and 7 denotes the highest trust. Though the participants do not know the robots, they are told that they will be cooperating with the same robot throughout the experiment. After eliminating broken cases, we get 175 unit trials for analysis.

We assess the capability of robots through the success rate and execution time. The task success score $s$ is defined as 1 if the target object is placed in the correct place, and 0 otherwise. The success rate (SR) is calculated by the ratio of success tasks. As a human-centric experiment, we use the
time-weighted success rate (TWSR) instead of path path-weighted success rate (PWSR). The TWSR score $s_t$ is calculated by the following formula:
\begin{align}
    s_t = s * \frac{T^*}{max(T^*, T)},
\end{align}
where $T$ denotes the time spent on the task, and $T^*$ is the minimum required time that is estimated by the distance of the shortest path. Besides, we record the adoption rate for robots to measure their help. A robot is considered adopted if it sends a message that is confirmed by the human. Participants can modify their adoption orally when mistouch the button, or exclude it from statistics for boundary cases.

\subsection{Results}

\subsubsection{Quantitative Analysis}
We present the quantitative scores in \Cref{tab: QuantResult}. Compared to the control group where no robot assistant exists, both the frontier robot and the oracle bring improvement in all quantitative metrics, including the success rate and the time-weighted success rate. As an approximation of the upper limit, the oracle improves the success rate significantly to 91.4\%, much more significant than the frontier agent which serves as a medium-level baseline. The oracle also has a much higher adoption rate, suggesting it is considered more helpful in collaboration. The significant discrepancies in quantitative metrics show that our system can successfully distinguish the capability of robots and thus be an eligible evaluation benchmark.

Another representative indicator, the average time spent across tasks is shown by the box plots in \Cref{fig: TimeSpent}. Although the robot successfully improves the overall performance, we find the improvement mainly appears in the hard tasks. As illustrated, the lower bound of time spent by the oracle and the frontier are almost the same, and the lower quartiles of all three settings do not have significant differences. This fits our insights that in household tasks, the robot plays its role when the task is at a specified level of difficulty. Too easy tasks in which the targets are very close to the robot or the human can not examine the capability of robots. As suggested by the results, our benchmark covers tasks of various levels of difficulty and thus can be used as a general evaluation platform. 

\subsubsection{Human trust}

We present the variation and average of human trust over trial times in \Cref{fig: HumanTrust}, where the shade denotes standard errors. Notably, the oracle always receives higher human trust than the frontier agent, which confirms that humans are aware of the distinction in robot capabilities and are willing to give higher trust to better agents. An interesting phenomenon is that since humans do not have the top view and are not aware of the robot's trajectory, they tend to measure the capability of robots from the time spent for searching. This results in the vibration of the trust scores, \eg, when the task is hard and the initial position of the target object is far from the robot, humans tend to reduce their trust scores due to the less help offered by the robot. Nevertheless, humans also consider the accumulated performance of the robot, reflected in the gradually converging trust scores. Besides, we find the trust scores for the oracle show a rising trend along with the progress of the experiment. We think the reason is that humans have noticed that the oracle can always find the target and raise the score, whereas the frontier agent still has a probability of failing.

The clear gap between the frontier agent and the oracle demonstrates that our experiment can successfully discriminate the capability of agents and serve as an effective benchmark for evaluating the power of agents and studying human trust in robots.

\begin{figure*}[t]
    \centering

    \includegraphics[width=\textwidth]{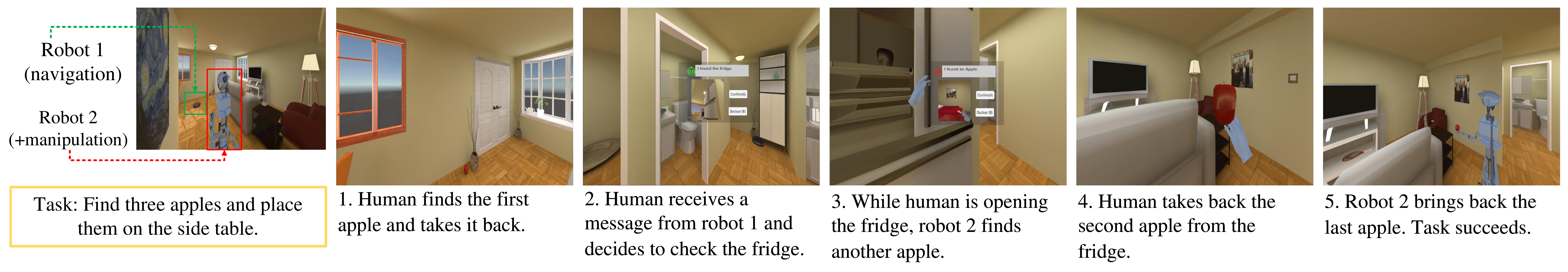}

    \caption{A demonstration of HRC in a multi-robot multi-target mobile manipulation task, where the human-robot team is asked to collect and place 3 apples. The complete task process is illustrated in the human's first perspective. Since the small robot can not interact with the fridge, it reports to the human and lets the human check the fridge. In contrast, the tall robot with manipulation capability can directly take the apple back.}
    \label{fig: multiPP_demo}
\end{figure*}

\subsection{Case study}

We present a case study to show the general collaboration process in \Cref{fig: CaseStudy}, in which the human is asked to find the remote and put it on the bed. The global maps show the layout of the house and the positions of task-related entities, where the human starts from the corridor and the robot is in the living room. 
As shown in trajectories, the oracle robot directly navigates to the target and is the first to report its position to the human. With such guidance, humans can quickly succeed with almost no extra exploration. Without prior knowledge, the frontier agent has to spend some time exploring and may even fail sometimes. Fortunately, it can still successfully find the target in many cases, resulting in saving some exploration time for humans. In contrast, when there is no robot assistant, the human may have to explore the full house and spend a lot of time. The human does not even find the target when the oracle-guided participant succeeds, while the frontier-assisted participant gets back to the bed with the remote. The comparison among the three trajectories demonstrates the assistance effect of different levels of robots on humans in collaboration tasks.
Since the picked case is of medium difficulty in a small house, all three shown trials finally succeed in nearly 75 seconds. However, the assistance of robots might be vital in hard tasks. For more cases and details, please refer to our accompanying video.

\begin{figure}[t]
    \centering
    \includegraphics[width=0.48\textwidth]{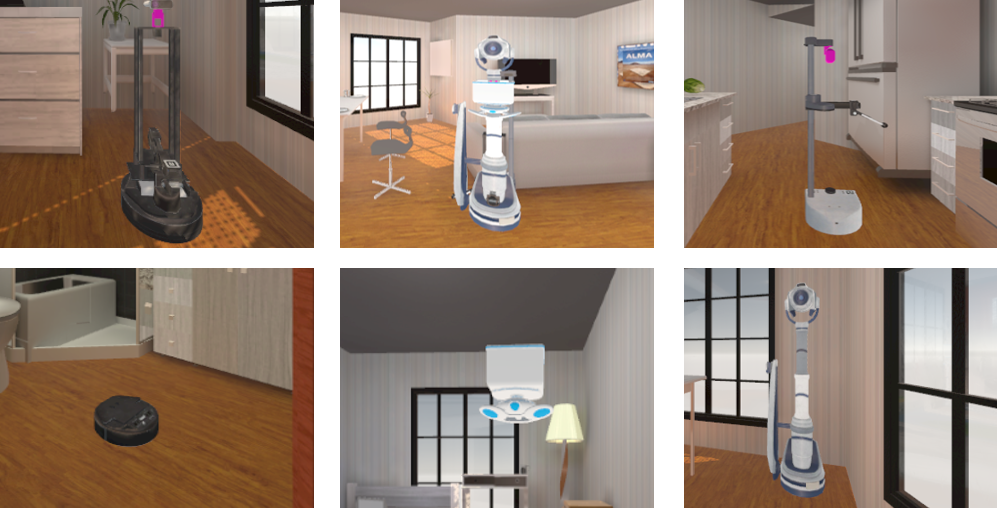}
    \caption{The HumanTHOR system supports various types of robots and is extensible for further customization. Current robots are from the ProcTHOR dataset.}
    \label{fig: Morerobots}
\end{figure}

\begin{figure}[t]
    \centering

    \includegraphics[width=0.48\textwidth]{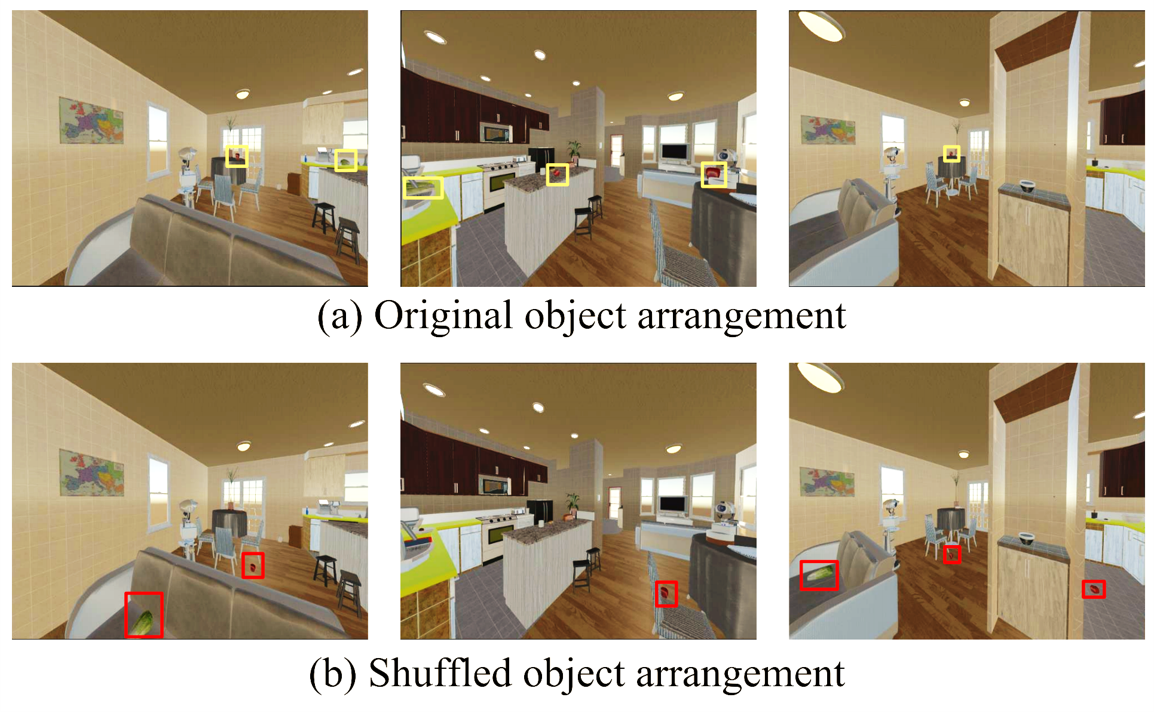}

    \caption{Comparison between the original scene and the shuffled one within three angles of view in the \textit{room rearrangement} task. The moved objects including \textit{apple}, \textit{lettuce}, and \textit{tomato}, are marked with gold squares in the original scene and with red squares in the shuffled scene.}
    \label{fig: rearrangement_demo}
\end{figure}

\begin{figure}[t]
    \centering

    \includegraphics[width=0.49\textwidth]{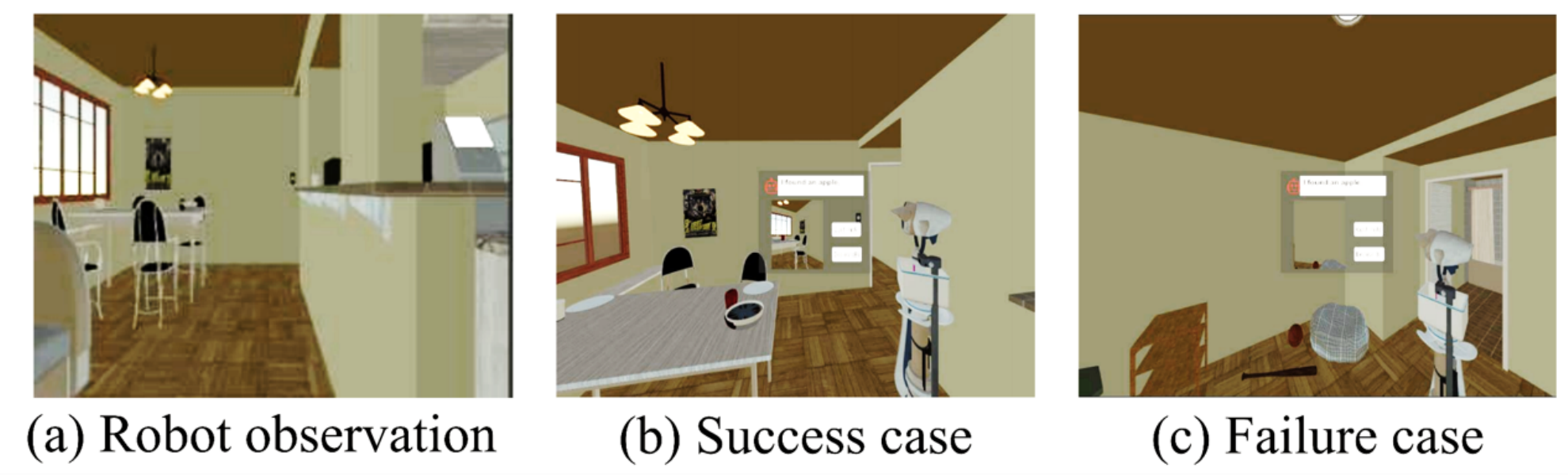}

    \caption{We present an example of employing the SAVN model to search for an apple in our HumanTHOR system. Success in simple cases demonstrates the practicability of leveraging deep learning models in our system, while failure in more cases suggests room for further study in robot algorithms. }
    \label{fig: savn_demo}
\end{figure}

\section{System Extensions}
\label{sec: Extension}

Based on the introduced functionalities, our platform is extensible for customized complex tasks and studies. 
In this section, we discuss the extensibility of HumanTHOR, including further support on the multi-robot setting, more complex tasks, and more robot algorithms. 

\subsection{Mutli-robot setting}
HumanTHOR supports incorporating multiple robots with the same or different types simultaneously in one scene.
To further exhibit this feature, we demonstrate a \textit{multi-robot multi-target mobile manipulation} task as illustrated in \Cref{fig: multiPP_demo}, in which the human has two heterogeneous robots as assistants. In this task, the taller robot has the manipulation capability, whereas the small robot can only navigate. Both of them can communicate with the human and contribute to the task in different ways.
The HumanTHOR system is also extensible for more types of robots, where the robots can have different appearance, size, performance, and capabilities. Currently available robots are illustrated in \Cref{fig: Morerobots}.

\subsection{More complex tasks}
As introduced in \Cref{sec: SystemAPI}, the HumanTHOR simulator exposes APIs for configuring scenes and objects on the fly, enabling multi-stage tasks such as \textit{room rearrangement} \cite{batra2020rearrangement}. In this task, the human and robots can first tour the clean environment and record the object arrangement, as shown in \Cref{fig: rearrangement_demo} (a). Once they are ready, some object arrangement will be shuffled as in \Cref{fig: rearrangement_demo} (b), requiring rearrangement.

Similarly, our system supports \textit{tidying up} \cite{sarch2022tidee}, another rearrangement task where no reference environment is provided and the agents need to rearrange objects according to commonsense, making the communication between the human and robots more important. For \textit{room rearrangement} and \textit{tidying up} tasks, we provide more details and corresponding demo videos with an oracle agent on our website.

\subsection{More robot algorithms}
As introduced in \Cref{sec: SystemAPI}, the HumanTHOR simulator provides ego-centric RGB observation and depth map. Therefore, vision-based navigation algorithms are also well supported. For demonstration, we reproduce a representative learning-based visual navigation model, SAVN \cite{wortsman2019learning}. As illustrated in \Cref{fig: savn_demo}, the existing visual navigation models are applicable in the HumanTHOR system. However, searching for objects in the whole house is still challenging and there exists room for further study.

\section{Conclusions}
\label{sec: Conclusions}
In this paper, we introduce HumanTHOR, an extended embodied simulator with an everyday task benchmark for studying human-robot collaboration in a shared workspace. Compared to existing environments and benchmarks, the HumanTHOR not only provides realistic simulation and a human-in-the-loop collaboration platform but is also scalable and flexible to support various collaborative robot studies such as human trust, emergent behavior, etc. 
Our preliminary user study results substantiate the helpfulness and importance of the robot assistant in human-robot collaboration by showing the improvement in overall quantitative performance and the gap indicates the room for further study on algorithms. Besides, the subjective results validate the availability of HumanTHOR in conducting human-related studies such as human trust. The quantitative and qualitative results have exhibited the effectiveness of our system in serving as a benchmark and experimental platform for HRC. 

Beyond the conducted experiments, the HumanTHOR system also supports advanced features such as multi-robot settings and more complicated tasks such as collaborative rearrangement. Customized robot algorithms such as learning-based visual navigation models are also well supported. In general, our HumanTHOR establishes a foundation and embodied test field for various future work in the human-robot collaboration domain.

\bibliographystyle{plainnat}
\bibliography{bib.bib}

\end{document}